# HPS: A Hierarchical Persian Stemming Method

Ayshe Rashidi[1] and Mina Zolfy Lighvan[2]

[1]Department of Electrical and Computer Engineering, Tabriz University, Tabriz, Iran
[2] Department of Electrical and Computer Engineering, Tabriz University, Tabriz, Iran

## ABSTRACT

*In this paper, a novel hierarchical Persian Stemming approach based on the Part-Of-Speech (POS) of the word in a sentence is presented. The implemented stemmer includes hash tables and several deterministic finite automata (DFA) in its different levels of hierarchy for removing the prefixes and suffixes of the words. We had two intentions in using hash tables in our method. The first one is that the DFA don't support some special words, so hash table can partly solve the addressed problem. And the second goal is to speed up the implemented stemmer with omitting the time that DFA need. Because of the hierarchical organization, this method is fast and flexible enough. Our experiments on test sets from Hamshahri Collection and Security News from ICTna.ir Site show that our method has the average accuracy of 95.37% which is even improved in using the method on a test set with common topics.*

## KEYWORDS

*Stemming, morphology, DFA machine, hash table, POS tags & hierarchical*

## 1. INTRODUCTION

Nowadays, people are surrounded by huge amount of information especially with the development of the internet. Hence, over the years many techniques are developed to help people manage and process their desired information. Many research themes in the field of artificial intelligence are emerging under this environment, for example, information retrieval, information extraction, information filtering, machine translation, question answering. Unfortunately, the words that seem in documents and in queries often have many morphological variants. In most cases, morphological variants of words have similar semantic interpretations and can be considered as equivalent for IR applications. Thus, pairs of terms such as "connect" and "connection" will not be recognized as equivalent without some form of natural language processing (NLP). So before the information retrieval from the documents the stemming techniques as an essential step are applied on the target data set to reduce the size of the data set which will improve the performance of IR System. So that a smaller data set or dictionary size results in a saving of storage space and processing time. There are several types of stemming algorithms which differ in respect to performance and accuracy. In this paper, we will describe some of them briefly and then also we will present our proposed method.

The organization of the rest of paper is as follows. Section 2, gives a brief background of Persian Language. Section three is a glance of related work. Section 4 describes our stemming method. In the Section 5, we test experimental results of our method, and Section 6 discusses our conclusion and suggestions.

International Journal on Natural Language Computing (IJNLC) Vol. 3, No.1, February 2014

## 2. RELATED WORKS

More frequently used stemming methods are: Affix removing, Look up Table and Statistics methods [1]. Affix Removing is depending on morphological structure of the language in which stemming is done by removing morphemes from any word. Porter algorithm is an example of this category[2], that is composed of 5 different steps. During these 5 steps more common affixes are removed using some special rules. Another example of this method is Krovatez [3], that uses a stemming procedure based on both inflectional and derivational suffixes in which the suffix stripping process is under the control of an English dictionary.

In the look up table based method, each word and its stem are stored in some look up tables, where for each stored word corresponding stem could be found. This method needs large storage space and its tables must be updated manually for each new word.

In the Statistics methods, using a process based on sets, rules are formulated according to the arrangement of words. n-gram [4], link analysis [5] and Hidden Markov Models [6] are examples of models that have been used in some statistics method for stemming.

In general, many works on stemming performance are reported in different fields for English language but not for other less popular language. For example for French language, Savoy [7] proposes a suffixing algorithm based on grammatical categories, also Savoy [8] presented another stemming procedure based on only a few general morphological rules. This approach corresponds to the English "S stemmer" method which conflates singular and plural word forms [9]. Tomlinson [10] evaluated the differences between Porter's stemmer [2] strategy and lexical stemmers (based on a dictionary of the corresponding language) for various European languages. For the Finnish and the German language, lexical stemmer tends to produce statistically better results, while for seven other languages performance differences were insignificant [11].

Two major algorithms for stemming in Persian language are presented. The first one has been proposed by Kazem taghva, Russell Beckley and Mohammad Sadeh in 2005 [12]. This method is an inspiration of the Porter algorithm in English [2], which is based on removing the suffix and prefix using Persian language morphology. For implementation of this method and to remove suffix and prefix from words, a DFA machine with 40 states is used. This method has some problems such as limited number of suffixes and low speed. The second algorithm is designed by GholamReza Ghasem Sani and Reza Hesamifard [13] which is based on the database or dictionary information of all the stems of the language. At first the input word should be searched in the database, if it is found, the stem will be returned, otherwise, the suffixes and prefixes should be removed and it should be searched again in database. Disadvantages of this method are its requirement to frequently database update, and high storage space.

## 3. PERSIAN LANGUAGE

The Persian language belongs to Indo-European languages, spoken and written primarily in Iran, Afghanistan, and a part of Tajikistan and is written using modified Arabic script, containing 28 Arabic letters and four more characters (ژ پ چ گ) __to express sounds not present in Classical Arabic and is a right to left language. In Persian, verbs involve tense, person, mode and its form (negative or positive). For example, the verb "می‌سازم" (mi-sazam: I make) is a present tense one consisting of three morphemes. "م" (am) is a suffix denoting first single person "ساز" (saz) is the present tense root of the verb and "می" (mi) is a prefix that expresses continuity.

Negative form of verbs is produced with adding "ن" (ne) to the first of them. For example, "نمی‌سازم" (ne-mi-saz-am - I don't make) is the negative form of the verb "می‌سازم" (misazam - I make). There are some certain rules to make verbs in Farsi language. A subset of these rules is shown in Table 1.

Table 1 Some morphological rules for verbs in Persian Language

| | | | |
|---|---|---|---|
| past tense (ماضی) | Simple (ساده) | بن ماضی + شناسه ماضی<br>past person identifier + past root | نوشتم(نوشت + م)<br>Neveştam = neweşt + am |
| | Continuous (استمراری) | می+بن ماضی+شناسه ماضی<br>past person identifier + past root + mi | می نوشتم(می+نوشت+م)<br>Mineveştam = mi + neveşt + am |
| | Present perfect (نقلی) | بن ماضی + 'ه' + شناسه ماضی نقلی<br>Present perfect past person identifier + 'h' + past root | نوشته ام(نوشت + ه + ام)<br>Neveşteam = Neveşt + e + am |
| | Unlikely (بعید) | بن ماضی + 'ه' + بود + شناسه ماضی<br>past person identifier + bud + 'h' + past root | نوشته بودم(نوشت + 'ه' + بود + 'م')<br>Neveşte budam = neveşt + e + bud + am |
| | Implicit (التزامی) | بن ماضی + 'ه' + باش + شناسه مضارع<br>present person identifier + baş + 'h' + past root | نوشته باشم(نوشت+ه+باش+م)<br>Neveşte başam = Neveşt + e + baş + am |
| Future tense (مستقبل) | -- | خواه + شناسه مضارع + بن ماضی<br>past root + Present person identifier + xãh | خواهم نوشت(خواه+م+نوشت)<br>Xaham neveşt= xãh+am+ neveşt |
| Present tense (مضارع) | Simple (ساده) | بن مضارع + شناسه مضارع<br>Present root + Present person identifier | نویسم(نویس + م)<br>Nevisam = Nevis + am |
| | Declarative (اخباری) | می+ بن مضارع + شناسه مضارع<br>Mi + Present root + Present person identifier | می نویسم(می + نویس + م)<br>Minevisam= mi + nevis + am |
| | Implicit (التزامی) | 'ب' + بن مضارع + شناسه مضارع<br>B + Present root + Present person identifier | بنویسم(ب+نویس+م)<br>Benevisam = be + nevis + am |
| | Imperative (امری) | 'ب'+ بن مضارع<br>B + Present root | بنویس(ب + نویس)<br>Benevis = be + nevis |

In Persian language we have a lot of rules for making nouns. In general, the plural forms of nouns are formed by adding the suffixes (ون، ین، ات، ان، ها). "ها" (hã) is used for all words. "ان" (ãn) is used for humans, animals and everything that is alive. Also, "ون، ین، ات" (ãt ,un , in) is used for some words borrowed from Arabic and some Persian words. There is another kind of plural form in Persian that is called Mokassar which is a derivational plural form (irregulars in Persian), that many of them borrowed from Arabic. Some examples of plural forms are shown in Table 2.

Table 2 Some Morphological Rules for Nouns in Persian Language

| Type | Suffixes | Word structure: Word= Word Stem + suffixes | |
|---|---|---|---|
| Plural | ان(ãn) | Deraxtãn=deraxt+ãn | درختان(trees) = درخت + ان |
| | ها(ha) | Dasthã=dast+hã | دست‌ها(hands) = دست + ها |
| | ات(ãt) | Nabãtãt=nabãt+ãt | نباتات(plants) = نبات + ات |
| | ین(in) ، ون(un) | Mo'alemun = mo'alem+ in | معلمین(teachers) = معلم + ین |
| Possession | ت(at)، م(am)، ش(aş) | Dastam=dast+am | دستم(my hand) = دست + م |
| | مان(mãn)، تان(tãn)، شان(ãşn) | Dastemãn=dast+mãn | دستمان(our hand) = دست + مان |
| Others | ی(i)، ه(h)، ک(k) | Xubi=xub+i | خوبی(goodness) = خوب + ی |
| | یت(yat)، چه(ĉe)، چی(ĉi) | Jam'yat=jam'+yat | جمعیت(population) = جمع + یت |
| | بان(bãn)، دان(dãn)، زار(zãr) | Bãghbãn=bãgh+bãn | باغبان(gardener) = باغ + بان |
| | واره(wãre) | Guşwãre=Guş+ware | گوشواره(eardrop) = گوش + واره |

There are some orthographic rules on the effects of joining affixes in some words. For example, consider a plural word consisting of two parts A and B. In such an example if the last letter of A and the first letter of B is "ا" (ã), a letter "ی" (y) is added between them. Assume A is "دانا" (dãnã - wise) and B is "ان" (ãn), the joining result is "دانایان" (dãnã-yãn: wises).

An adjective is a word or group of words that appears before or after a noun, and explains a feature or concept about it. Adjectives have different types such as simple, nominative, participle, relative and merit. Here, we categorized them based on the number of suffixes letters, because our method is based on morphology. Some of common types of adjectives are presented in table 3.

Table 3 Some Morphological Rules for Adjectives in Persian Language

| | Suffixes | Word structure: Word= Word Stem + suffixes | |
|---|---|---|---|
| Adjective | ا(ã)، ی(i)، ه(h) | Dãrã = dãr + ã | دارا(wealthy) = دار + ا |
| | تر(tar)، گر(gar)، ین(in)، ار(ãr) | Xubtar = xub + tar | خوبتر(Better) = خوب + تر |
| | انه(ãne)، مند(mand)، ناک(nãk)، وار(wãr) | Mahramãneh = mahram + ãne | محرمانه(Confidential) = محرم+ انه |
| | ترین(tarin)، گانه(gãne) | Xubtarin = xub + tarin | خوبترین(best) = خوب + ترین |

Similar to the nouns, there are some orthographic rules for adjectives in Persian language. For example if we want to make a relative adjective from a word(with adding 'ی'(i) to end of it) that has a 'ه' (h) as its last letter like "بانه"( Baneh: a city name), we should add an "ا" (a) letter between them so relative adjective for "بانه" is "بانه‌ای"(Baneai: from Bane).

## 4. HPS METHOD

### 4.1. Description of our HPS method

For stemming a textual document or a sentence, an effective stemming method should focus mainly on nouns, adjectives and verbs because these words carry the major meaning of a sentence or a document. Therefore, in this paper we ignore the stemming of other components of sentence. Persian language as well as English language has affixation morphology, means that for the different applications or to create the new meaning of a word, prefix and suffix stick to the begin and end of the words. Persian nouns as well as English nouns have plural and ownership suffixes. Persian verbs according to tense, person, negative and modes are different and have more variety than English verbs. Also Persian has so many adjective suffixes.

HPS (hierarchical Persian Stemmer) method employs a hierarchical process based on morphology and POS tags. It has three distinct parts for nouns, adjectives and verbs suffix stemming. In addition HPS uses hash table for stemming of some exceptions that other stemmer can't support it.

In HPS the stemming task is spread into several hierarchical levels. Figure 1 shows a Block Diagram of different levels of HPS method. The first level of HPS is showed by PreStemmer-DFA which is responsible of removing prefixes from the words. The Next level named SufStemmer removes suffixes and is composed of three distinct parts based on the POS tags (N for nouns, V for verbs and A for adjectives). Each of the mentioned parts contains of two levels that composed by a hash table and a DFA. For example in the first part that is belong to the nouns, N_Hash is a hash table that constructs the first level and SufStemmer_NDFA is the DFA based stemmer of the corresponding second level.

HPS method stores some particular words like high frequency words , Mokassar plural words that borrowed from Arabic and irregular plurals and some words like "سازمان"(sãzemãn: organization) in three distinct small hash tables(N_Hash for nouns, A_Hash for adjectives and V_Hash for verbs). In the diagram of the method NFile, AFile and VFile are files that containing noun, adjectives and verbs words respectively those stores in corresponding hash tables.

Our stemmer uses a lower bound limit on stem length (which is equal to three here) and it also follows some rules on the last letter of words and the first letters of suffixes. HPS at first identifies prefixes, and removes prefix according to defined sequences in the existence paths in the PreStemmer-DFA.

We have grouped suffixes into three main groups as verb-suffixes (VL1, VL2, VL4, VL5, VL6, VL7), noun-suffixes (Pl2, Plo3, Po1, Po2, Po3, Ot1, Ot2, Ot3, Ot4), and adjective-suffixes (AL1, AL2, AL3, AL4) and each of this main groups has sub groups based on number of suffix letters (and type of suffix for the noun-suffixes). This grouping indicates the number of suffix letters that would be cut from the word. If the stemmer first identifies the prefix "ن" (n) in the word "ننوشتیم" (naneveştim: we did not write) as a prefix, it then identifies suffix "یم" (yam) and removes it to produce the stem "نوشت" (neveşt: wrote).

Noun suffixes are stacked according to this pattern (reading from right-to-left):

(Possessive) + (Plural) + (Other) < Stem >

For example, the stemmer first finds the possessive noun suffix "یمان"(yemãn) in the word "نوشته‌هایمان" (neveştehãyemãn: our writings"), then it finds the plural noun suffix " ها"(hã) and, finally, it finds the other-noun-suffix "ه"(h) to reach the stem " نوشت " (neveşt: wrote). Hence the stemmer removes up to three suffixes from nouns.

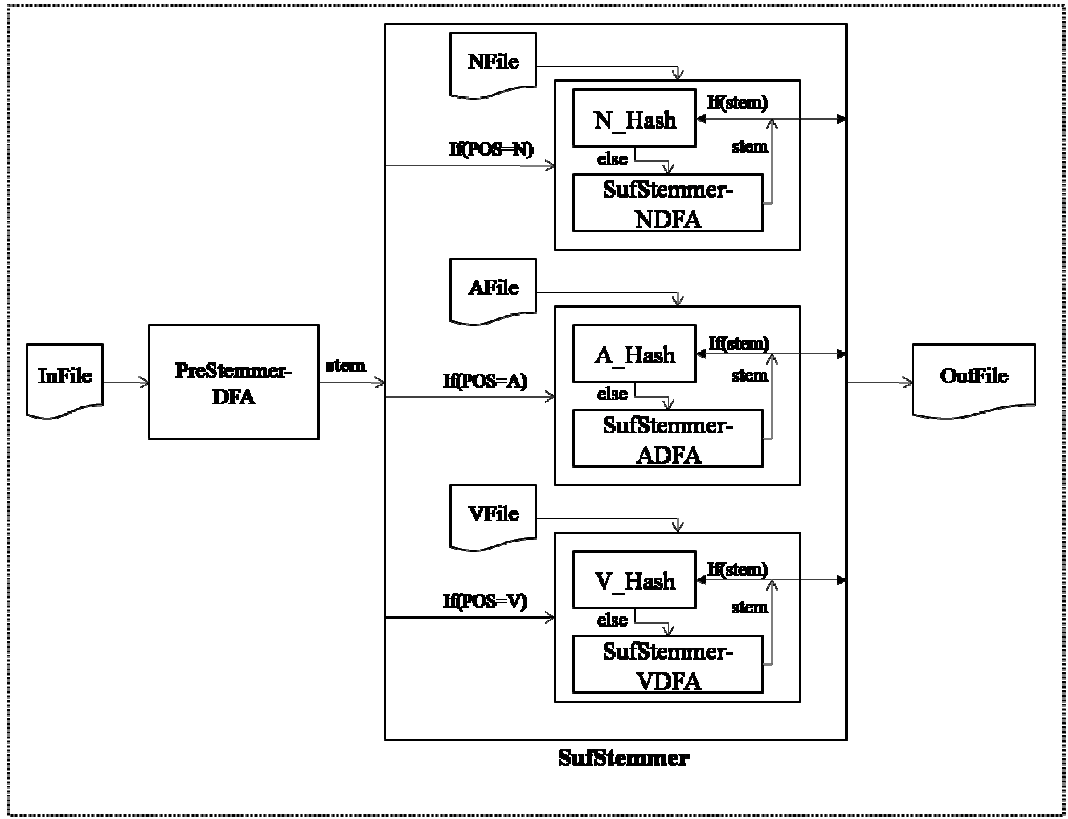

Figure 1 Diagram of HPS Proposed Method

## 4.2. Implementation

We implemented our proposed HPS method with a composition of three hash tables and four DFA (deterministic finite automata) machines. The hash tables are belonged to three major parts of word stemmer as described before. One of the four implemented DFA machines takes the role of prefix stemmer and the other three are for removing the suffixes from the words based on POS tags (noun: N, adjective: A or verb: V).

The prefix DFA stemmer runs on the input word and if detects a prefix pattern then removes it. Depending on the POS tag of a word its corresponding hash tables is being searched, in the case of finding the word in the hash table, related stem is returned otherwise corresponding suffix DFA stemmer is being run to remove the suffixes during the states of the DFA. If the generated word is a stem then the process is completed otherwise it will be returned again to the hash table.
It is remarkable that a word may have multiple suffixes, so for removing all suffixes, output will be given back to the suffix stemmer system as a new word and this process repeated until it can't find any more suffix or returned word is contained less than three letters.

Depends on POS of input word, a small array for storing suffix groups is used. We have named all existence states in the DFAs, as "NIL" or one of suffix groups in the suffix DFA stemmers and

16

"NIL" or "PRE" in the Prefix DFA stemmer. If final state was "NIL" then not removed any things from the input word means that the input word is returned as its stem, else regard to the suffix group of final state, related suffix will be removed. Figure 2 shows a simple DFA machine which has been used for removing two noun suffixes subsets: Plo3= {"شان","تان","مان"} and Pl2={"ین","ون","ات","ها","ان"}. The three groups of states of this DFA are showed in Table 4. For example, consider "کیفشان" (kifeşān= their bag) as an input word. The DFA gets the words from left to right that means the last letter of the word ('ن') is the first one the DFA gets. Therefore applying the example word ("کیفشان") will terminate in state 9 that is grouped as "Plo3". Thus three letters of "شان" (şān) suffix will be cut from the end of input word and "کیف" (kif) has been returned as the stem.

Table 4 An example for grouping of the final state

| Final States | Suffix group |
|---|---|
| 1,2,3,4 | NIL |
| 5,6,7,11,12 | Pl2 |
| 8,9,10 | Plo3 |

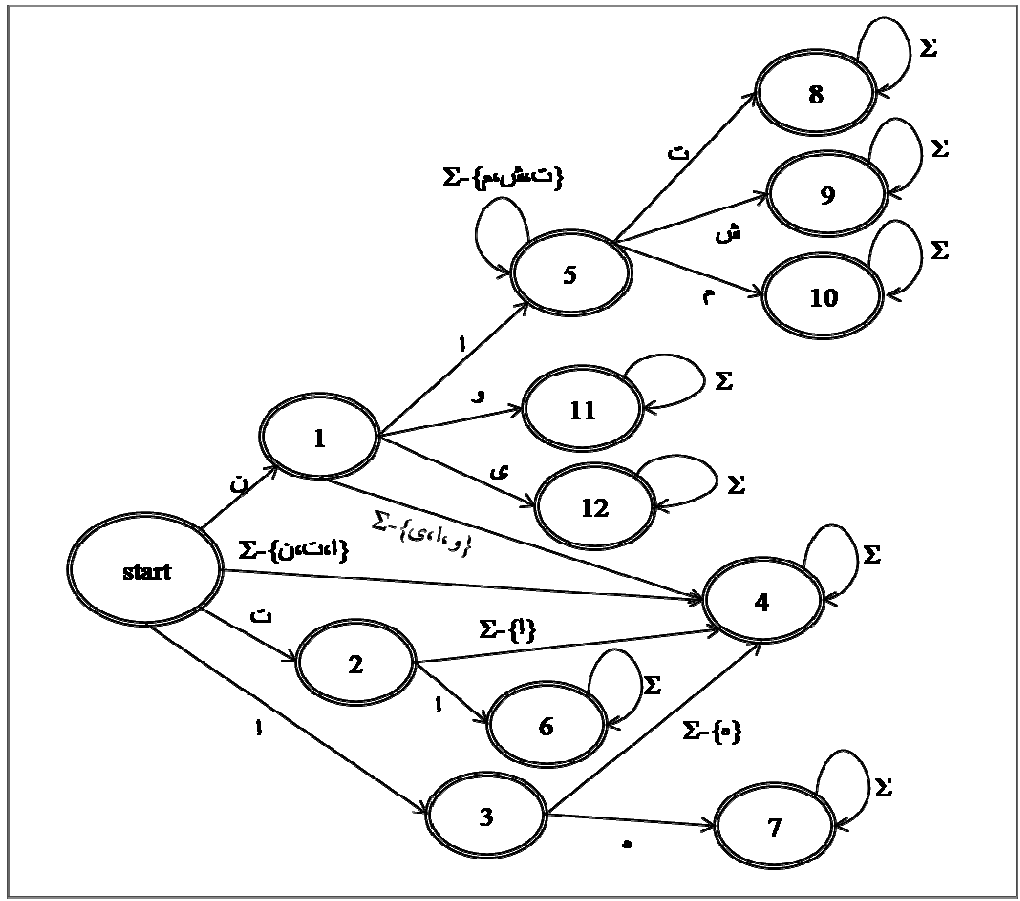

Figure 2 An example of a Small DFA machine

## 5. EXPERIMENTAL RESULTS

For evaluating the proposed HPS method on the Persian language, *Hamshahri Collection* (with various topics) and *Security News from the ISTna.ir site*(with the special security topic) have been used, so that we create some different test sets with different sizes, then we test the HPS algorithm on each of them. The creation of test sets is as follows: first, we select some test documents with different lengths (small to large) from the two mentioned corpus, and then give them to a POS (Part Of Speech) tagger system like [14] for detecting POS of all words of documents. Then, we hold only words that have Noun, Adjective or Verb POS tags and stored these words and their POS tags in the two distinct files as the inputs of our system. We assumed that nouns, adjectives and verbs are most meaningful parts of sentences of texts, therefore remaining components of sentences such as adverbs, conjunctions, determiner, number, propositions, pronouns and punctuations are ignored. The results that are shown in Table 5, Table 6 have relatively good accuracy. Most of the incorrect results are related to compound words because of many of them haven't specified morphology rules.

Table 5 Test of HPS method on the Hamshahri Collection

| Test set No. | Topic | Words ( noun, adjective and verb) | Correct Results | Wrong Results | Accuracy (%) |
|---|---|---|---|---|---|
| 1 | Literature & Art | 24 | 23 | 1 | 95.8 |
| 2 | Literature & Art | 48 | 45 | 3 | 93.7 |
| 3 | Literature & Art | 72 | 67 | 5 | 93.1 |
| 4 | Literature & Art | 99 | 92 | 7 | 93 |
| 5 | Literature & Art | 150 | 140 | 10 | 93.3 |
| 6 | Literature & Art | 247 | 234 | 13 | 94.7 |
| 7 | social | 117 | 113 | 4 | 96.5 |
| 8 | social | 324 | 314 | 10 | 96.9 |
| 9 | science & culture | 131 | 127 | 4 | 96.9 |
| 10 | science & culture | 246 | 240 | 6 | 97.5 |
| 11 | science & culture | 394 | 385 | 9 | 97.7 |
| **Average of Accuracy = 95.37** | | | | | |

Table 5 shows the experimental results of applying HPS on a test set composed of texts on different topics from the *Hamshahri Collection*. The Correct Results column indicates number of words stemmed correctly and the Wrong Results indicates number of incorrectly stemmed words plus not stemmed words. The Accuracy is the percentage of correct results between all words. The average accuracy of 95.37% is a reasonable result which shows the performance of HPS method.

Another experiment has been done on a test set composed of texts with common topic on security and the results are showed in Table 6. In this table the stemming results of using hash tables are compared to the results of not using them. Obtained results shows that hash tables have remarkable influence on the stemming accuracy which has increased it by 4%.

Table 6 Test of HPS method on *Security News* from *ICTna.ir* (with hash table and without hash table)

| Text No. | Word No. | With Hash Table | | | Without Hash Table | | |
|---|---|---|---|---|---|---|---|
| | | Correct | Wrong | Accuracy (%) | Correct | Wrong | Accuracy (%) |
| 1 | 72  | 68  | 4  | 94.44 | 62  | 10 | 86.11 |
| 2 | 94  | 91  | 3  | 96.80 | 89  | 5  | 94.68 |
| 3 | 188 | 182 | 6  | 96.80 | 176 | 12 | 93.61 |
| 4 | 211 | 199 | 12 | 94.31 | 190 | 36 | 90.04 |
| 5 | 214 | 203 | 11 | 94.85 | 196 | 34 | 91.58 |
| 6 | 215 | 210 | 5  | 97.67 | 199 | 21 | 92.55 |
| 7 | 349 | 331 | 18 | 94.84 | 320 | 29 | 91.69 |
| 8 | 179 | 170 | 9  | 94.97 | 164 | 14 | 91.62 |
| | | **Average of Accuracy = 95.58** | | | **Average of Accuracy = 91.45** | | |

## 6. CONCLUSIONS

In this paper the HPS methods for Persian stemming is presented. The novelty of this method is because of its hierarchical structure which is composed of different levels based on DFAs and hash tables. Using DFAs and hash tables together provides taking advantages of both of them.
In HPS the words are categorized based on their POS tags which reduce the probability of mistaken results. The structured design of HPS made this method dynamic and extensible. Using individual DFAs for the words with different POS tags increased the speed of stemming and also made it more extensible.

The main goal in introducing HPS was stemming on the texts with special topics therefore we have used small hash tables of the words on special topics. This idea increases the accuracy of stemming and also increases the stemming task speed because searching in small hash table is fast enough and also the words found in hash tables don't go through DFAs.

The experimental result shows the average accuracy of 95.37% which is even improved in using the method on a test set with common topics. Comparing the results with the similar works such as [12, 13, 15] shows the advantages of HPS method.

**Ayshe Rashidi** received the B.S.c degree in Computer Engineering (Hardware) from Technical and Engineering faculty, Shahed University, Tehran, Iran in 2011. She is currently M.Sc. student in Computer Engineering (Artificial Intelligent) from Electrical and Computer Engineering faculty of Tabriz University, Iran. Her research interests include Algorithm Design, Data Mining, Text Processing, NLP, and   Intrusion Detection Systems, Information Extraction and Retrieval.

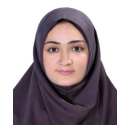

**Mina Zolfy Lighvan** received the B.Sc degree in Computer Engineering (hardware) and M.Sc. degree in Computer Engineering (Computer Architecture) from ECE faculty, university of Tehran, Iran in 1999, 2002 respectively. She received Ph.D. degree in Electronic Engineering (Digital Electronic) from Electrical and Computer Engineering faculty of Tabriz University, Iran. She currently is an assistant professor and works as a lecturer in Tabriz university. She has more than 20 papers that were published in different national and international conferences and Journals. Dr. Zolfy major research interests include Text Retrieval, Object oriented Programming & Design, Algorithms Analysis, HDL Simulation, HDL Verification, HDL Fault Simulation, HDL Test Tool VHDL, Verilog,  hardware test, CAD Tool, synthesis, Digital circuit design & simulation.

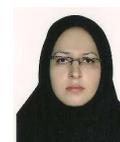